\documentclass[runningheads]{llncs}
\usepackage[T1]{fontenc}
\usepackage{amsmath,graphicx}
\usepackage{amsmath,amsfonts,bm}

\def\eqref#1{equation~\ref{#1}}
\def\1{\bm{1}}

\def\va{{\bm{a}}}

\def\vh{{\bm{h}}}

\def\vp{{\bm{p}}}

\def\vx{{\bm{x}}}

\def\mA{{\bm{A}}}

\def\mW{{\bm{W}}}
\def\mX{{\bm{X}}}

\DeclareMathAlphabet{\mathsfit}{\encodingdefault}{\sfdefault}{m}{sl}
\SetMathAlphabet{\mathsfit}{bold}{\encodingdefault}{\sfdefault}{bx}{n}

\def\gN{{\mathcal{N}}}

\DeclareMathOperator*{\argmin}{arg\,min}

\newcommand{\eg}{e.\,g., }
\newcommand{\ie}{i.\,e., }
\newcommand{\neighborhood}[1]{\gN({#1})}
\usepackage{adjustbox}
\usepackage{amsmath}

\newcommand{\closedneighborhood}[1]{\overline{\gN}({#1})}
\usepackage{multirow}
\usepackage{booktabs}

\usepackage{mathtools}
\usepackage{lipsum}% http://ctan.org/pkg/lipsum
\usepackage[]{graphicx}% http://ctan.org/pkg/graphicx
\usepackage{amsmath}
\usepackage{xcolor}

\usepackage{subcaption}

\usepackage{graphicx}
\begin{document}
\title{Supervised Attention Using Homophily in Graph Neural Networks}
\author{Michail Chatzianastasis  \and
Giannis Nikolentzos \and
Michalis Vazirgiannis }
\authorrunning{Michail Chatzianastasis}
\institute{LIX, École Polytechnique, IP Paris, Palaiseau, France}
\maketitle              % typeset the header of the contribution
\begin{abstract}
Graph neural networks have become the standard approach for dealing with learning problems on graphs.
Among the different variants of graph neural networks, graph attention networks (GATs) have been applied with great success to different tasks.
In the GAT model, each node assigns an importance score to its neighbors using an attention mechanism. 
However, similar to other graph neural networks, GATs aggregate messages from nodes that belong to different classes, and therefore produce node representations that are not well separated with respect to the different classes, which might hurt their performance. 
In this work, to alleviate this problem, we propose a new technique that can be incorporated into any graph attention model to encourage higher attention scores between nodes that share the same class label.
We evaluate the proposed method on several node classification datasets demonstrating increased performance over standard baseline models. 
\end{abstract}
\section{Introduction}
\label{sec:intro}

Graph neural networks (GNNs) have recently emerged as a general framework for learning graph representations and have been applied with great success in different domains such as in bioinformatics~\cite{li2022graph}, in physics~\cite{NEURIPS2020_c9f2f917} and in natural language processing~\cite{nikolentzos2020message}, just to name a few. Among others, GNNs have been used to generate molecules with specific chemical characteristics~\cite{mahmood2021masked}, to predict compound–protein interactions for drug discovery~\cite{tsubaki2019compound} and to detect misinformation in social media~\cite{han2020graph}.

While different types of GNNs have been proposed, most of these models follow an iterative message passing scheme, where each node aggregates information from its neighbors~\cite{gilmer2017neural}.
One of the most popular classes of this kind of models are the graph attention networks (GATs)~\cite{velicko2018,brody2022how,chatzianastasis2023graph,kim2021how}.
GATs employ an attention mechanism which can capture the importance of each neighbor and are thus considered state-of-the-art models in various graph learning tasks.
These models are also highly interpretable since the learned attention scores can provide information about the relevance of the neighboring nodes.

Unfortunately, real-world graphs often contain noise, as there usually exist edges between unrelated nodes.
In such a setting, once multiple message passing steps are performed, nodes will end up receiving too much noisy information from nodes that belong to different classes, thus leading to indistinguishable representations.
This problem is known as oversmoothing in the graph representation learning literature~\cite{chen2020measuring,cai2020note}, and can dramatically harm the performance of GNNs in the node classification task.
Several approaches have been proposed to address the issue of oversmoothing such as normalization layers~\cite{Zhao2020PairNorm,dasoulas2021lipschitz}, generalized band-pass filtering operations~\cite{min2020scattering}, and approaches that change the graph structure~\cite{chen2020measuring}.
However, most of them are computationally expensive or require extensive architectural modifications.

In this work, we focus on removing the noisy information from the graph using an attention mechanism.
Specifically, we propose a new loss function that encourages nodes to mainly attend to nodes that belong to the same class, and to a lesser extent to nodes that belong to different classes by supervising the attention scores.
To motivate our approach, we first experimentally verify that GNNs perform better as the edge homophily in the graph increases, \ie as we remove inter-class edges. Therefore, it is important to learn attention scores close to $0$ for the inter-class edges.
Furthermore, we demonstrate that the proposed method outperforms various baselines in real-world node classification tasks.
Our approach is computationally efficient, and it can be applied to any graph attention model with minimal modifications in the architecture.
Finally, we visualize the distribution of the learned attention scores of our proposed model and of vanilla graph attention networks, for the intra- and the inter-class edges.
We verify that our proposed model learns higher attention scores for the intra-class edges, leading to high quality node representations.
%Our approach is computationally efficient and can be applied to any graph attention model with minimal modifications in the architecture.

Our contributions can be summarized as follows: \begin{itemize}
\item We show experimentally that GNNs perform better as the edge homophily in the graph increases, and that it is important to learn attention scores close to 0 for the inter-class edges.

 \item We propose a novel loss function for attentional GNNs that encourages nodes to attend mainly to nodes that belong to the same class, and to a lesser extent to nodes that belong to different classes.

\item We show that our approach outperforms various baselines in real-world node classification tasks.

\end{itemize}
The rest of the paper is organized as follows. Section \ref{sec:related_work} presents the related work. 
Section \ref{sec:method} introduces the proposed loss function. Finally, Sections \ref{sec:exp} and \ref{sec:conclusions} present the experimental results and conclusions, respectively.

\section{Related Work}
\label{sec:related_work}
Graph Neural Networks (GNNs) have received significant attention in the past years, with a growing number of research works proposing novel methods and applications.
The first GNN models were proposed several years ago~\cite{sperduti1997supervised,scarselli2008graph}, however with the rise of deep learning, GNNs have gained renewed interest in the research community~\cite{kipf2017semi,hamilton2017inductive}.
The majority of GNN models can be reformulated into a single common framework known as Message Passing Neural Networks (MPNNs)~\cite{gilmer2017neural}.
These models iteratively update a given node's representation by aggregating the feature vectors of its neighbors.
Graph attention networks (GATs) correspond to one of the major subclasses of MPNNs~\cite{velicko2018,brody2022how,chatzianastasis2023graph}.
GATs employ an attention mechanism which allows them to incorporate explicit weights for each neighbor.
One of the main advantages of these models is that they are highly interpretable due to the learned attention scores.
Numerous studies have proposed several enhancements and expansions to the message passing mechanism of MPNNs.
These include among others, works that use more expressive or learnable aggregation functions~\cite{murphy2019relational,seo2019discriminative,dasoulas2021learning,chatzianastasis2023graph}, schemes that operate on high-order neighborhoods of nodes~\cite{abu2019mixhop,jin2020gralsp,nikolentzos2020k}, and approaches that operate in the hyperbolic space~\cite{liu2019hyperbolic,chami2019hyperbolic,nikolentzos2023weisfeiler}.
However, a common issue that affects the performance of various MPNNs is oversmoothing.
Several studies have investigated the causes and effects of oversmoothing, as well as potential solutions to mitigate this problem, including normalization techniques~\cite{yang2020revisiting,dasoulas2021lipschitz} and graph rewiring methods~\cite{chen2020measuring,karhadkar2022fosr}.

\section{Methodology}\label{sec:method}

\subsection{Preliminaries}
Let $G=(V,E)$ be an undirected graph where $V$ is a set of nodes and $E$ is a set of edges.
We will denote by $N$ the number of vertices and by $M$ the number of edges, \ie $N = |V|$ and $M = |E|$.
Then, we have that $V = \{ v_1, v_2, \ldots, v_N\}$.
Let $\mA \in \mathbb{R}^{N \times N}$ denote the adjacency matrix of $G$, $\mX=[\vx_1,\vx_2,\ldots,\vx_N]^\top \in \mathbb{R}^{N \times d}$ be the matrix that stores the node features, and $\bm{Y} = [y_1,y_2,\ldots,y_N]^\top \in \{1,\ldots,C\}^{N}$ the vector that stores the nodes' class labels where $C$ is the number of classes. 
Let $\mathcal{N}(i)$ denote the indices of the neighbors of node $v_i$, \ie the set $\{j \colon \{v_i,v_j\} \in E\}$.
We denote the features of the neighbors of a node $v_i$ by the multiset $\bm{X}_{\neighborhood{i}} = \{\vx_j \colon j \in \neighborhood{i}\}$.
We also define the neighborhood of $v_i$ including $v_i$ as $\closedneighborhood{i}= \neighborhood{i}\cup \{ i\}$ and the corresponding features as $\bm{X}_{\closedneighborhood{i}}$.
Given a training set of nodes, the goal of supervised node classification is to learn a mapping from the node set to the set of labels, $f: V \rightarrow \{0,1,\ldots,C\}$.

\subsection{Graph Neural Networks}
GNNs typically use the graph structure $\bm{A}$ along with the node features $\bm{X}$ to learn a representation $\vh_i$ for each node $v_i \in V$~\cite{gori2005new}. 
As already discussed, most GNNs employ a message-passing scheme~\cite{gilmer2017neural} where every node updates its representation by aggregating the representations of its neighbors and combining them with its own representation.
Since there is no natural ordering of the neighbors of a node, the aggregation function needs to be permutation invariant, and usually has a significant impact on the performance and the expressiveness of the GNN model~\cite{xu2019powerful}.
Common aggregation functions include the sum, mean, max, and min operators, but also attention-based pooling aggregators~\cite{kipf2017semi,hamilton2017inductive}.

In this work, we mainly focus on attention-based aggregators, where the representation of each node $v_i$ is updated using a weighted sum of the representations of its neighbors:
\begin{equation}
    \vh_i = \sigma \left( \sum_{j \in \closedneighborhood{i}} \alpha_{ij} \bm{W}\vh_j \right)
    \label{eq:mp}
\end{equation}
where $\vh_i \in \mathbb{R}^d$ denotes the hidden representation of node $v_i$, $\bm{W} \in \mathbb{R}^{d_o \times d}$ is a weight matrix and $\alpha_{ij}$ is the learned attention score (\eg how much node $v_i$ attends to node $v_j$).
Equation~(\ref{eq:mp}) is applied iteratively, however, for ease of notation, we have dropped the superscript (that denotes the iteration number).
Among the different attention models that have been proposed in the past years, the Graph Attention Network (GAT)~\cite{velicko2018} computes the attention scores by applying a single-layer feedforward neural network in the concatenated node features of the two nodes, while GATv2~\cite{brody2022how}, an improved version of GAT, computes more expressive and dynamic attention. 
In our experiments, we use GATv2 as the backbone network, but our approach can be easily applied to any graph attention model. 
Specifically, we compute the un-normalized attention score between two nodes $v_i, v_j$ using the following equation:
\begin{equation}
 e_{ij} =  
\va
^{\top}
	\mathrm{LeakyReLU}
	\left(
		\mW_2 \left[\vh_{i} \| \vh_{j}\right] 
	\right) \label{eq:gat2}
\end{equation}
where $\va \in \mathbb{R}^{d_o}$ is a weight vector and $\mW_2 \in \mathbb{R}^{d_o \times 2d}$ a weight matrix. 
Then, we apply the softmax function to normalize the attention scores across all neighbors of $v_i$:
\begin{equation}
	\alpha_{ij} =
	\frac{\mathrm{exp}\left(e_{ij}\right)}{\sum\nolimits_{k\in\mathcal{N}(i)} \mathrm{exp}\left(e_{ik}\right)}
	\label{eq:softmax}
\end{equation}

%\begin{equation}
%	\alpha_{ij} = \frac{\exp\left(\text{LeakyReLU}\left(\vec{\bf a}^T[{\bf %W}\vec{h}_i\|{\bf W}\vec{h}_j]\right)\right)}{\sum_{k\in\mathcal{N}_i} %\exp\left(\text{LeakyReLU}\left(\vec{\bf a}^T[{\bf W}\vec{h}_i\|{\bf %W}\vec{h}_k]\right)\right)}
%\end{equation}

\subsection{Problem of Information Mixing}

\begin{definition}[Edge Homophily]{Given a graph $G = (V, E)$ with a vector of node class labels $\bm{Y}$, the edge homophily ratio is the fraction of intra-class edges in the graph, \ie the fraction of edges that connect nodes with the same labels.}

\begin{equation}
h(G,\bm{Y}) = \dfrac{\big|\big\{(v_i,v_j) \colon (v_i, v_j) \in E \land y_i=y_j \big\}\big|}{|E|} 
\end{equation}
\end{definition}

The problem of information mixing or oversmoothing~\cite{chen2020measuring,wang2019improving} occurs mainly in cases where there are edges between nodes that belong to different classes (\ie inter-class edges).
In each message passing iteration, information will be exchanged through these ``noisy'' edges, leading nodes that belong to different classes into obtaining highly non-separable representations.
Therefore, the node classification task is becoming extremely challenging.
Ideally, we would like to identify and eliminate those ``noisy'' edges, such that nodes will only aggregate information from intra-class edges. 

In this paper, we leverage graph attention networks in order to alleviate this issue.
Specifically, we encourage the network to learn attention scores that minimize information mixing in the graph.
Note that a node $v_i$ receives noisy information as follows:
\begin{equation}
\text{noise}(i) = \sum_{j \in \mathcal{N}_{\text{inter}}(i)} \alpha_{ij}\bm{W}\bm{h}_j
\end{equation}
where $\mathcal{N}_{\text{inter}}(i)$ is the set of indices of the inter-class neighbors of node $v_i$.
Therefore, we would like our attention scores to satisfy the following equation:
\begin{equation}
    \Big\{ \alpha_{ij}^* \colon j \in \mathcal{N}_{\text{inter}}(i) \Big\} = \argmin_{ \big\{ \alpha_{ij} \colon j \in \mathcal{N}_{\text{inter}}(i) \big\} } \sum_{j \in \mathcal{N}_{\text{inter}}(i)} ||\alpha_{ij}\bm{W}\bm{h}_j||
\end{equation}
The solution of the above equation gives us $\alpha_{ij}=0$ for all the inter-class edges $\{v_i,v_j\}$.

\begin{figure}[t]
    \centering
    \includegraphics[width=\textwidth]{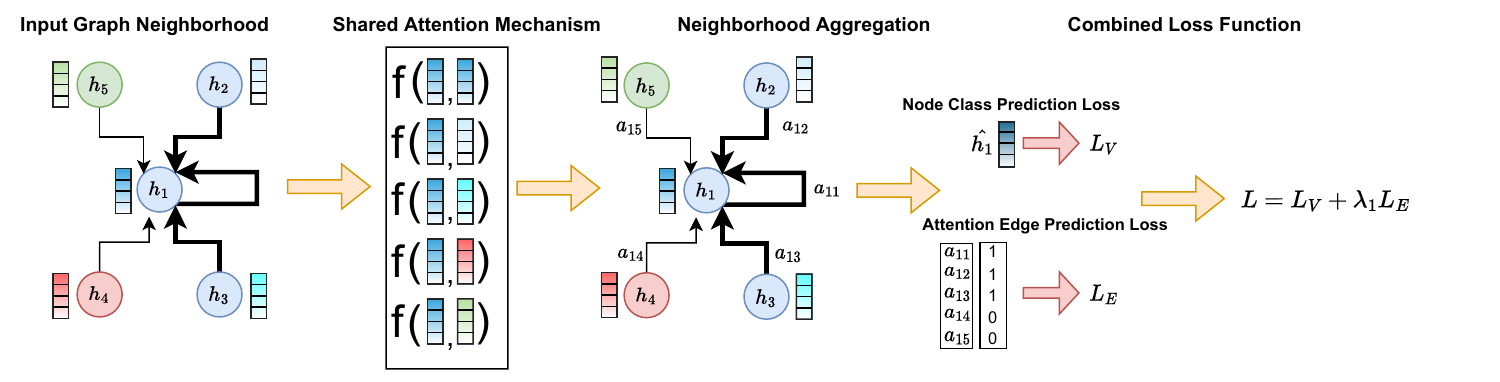}
    \caption{An illustration of our proposed method (\textbf{HS-GATv2}). We combine the loss $L_V$ from the node classification task and the loss $L_E$ from the attention scores based on the training edges, to push neighbor nodes with the same class to have large attention scores and nodes from different classes to have small attention scores. Our approach is applicable to any graph attention model, by setting accordingly the attention function $\textbf{f}$.}
    \label{fig:architecture}
\end{figure} 

\subsection{Supervised Attention using Homophily (HS-GATv2)}

% add equation of minimizing information mixing. 

Based on the previous analysis, we propose a new loss function for training graph attention networks that deals with the information mixing problem.
Specifically, we propose to supervise the attention scores between the edges, by providing labels that indicate if the edge is an intra- or an inter-class edge.
Let $V_{\text{train}}$ denote a set that contains the indices of the nodes that belong to the training set.
Let also $E_{\text{train}}$ denote the training edge set which consists of all the edges where both source and target nodes belong to the training node set, \ie $E_{\text{train}} = \big\{ \{v_i,v_j\} \colon \{v_i,v_j\} \in E \land i \in V_{\text{train}} \land j \in V_{\text{train}} \big\}$.
Formally, the proposed loss function combines the following two terms: (1) the cross-entropy loss between model predictions and class labels of nodes (denoted by $L_V$); and (2) the supervised attention losses for the edges between nodes of the training set (denoted by $L_E$) with mixing coefficient $\lambda$:
\begin{equation}
    \begin{split}
        L &= L_V + \lambda \, L_E \\
        L_V &= - \frac{1}{|V_{\text{train}}|} \sum_{i \in V_{\text{train}}} \sum_{\mathclap{c=1}}^C y_{i,c}\log(\vp_{i,c}) \\
        L_E &= -\frac{1}{T \, |E_{\text{train}}|} \sum\limits_{t=1}^T \sum_{e\in E_{\text{train}}} \Big(y_e \log \big(\sigma(e_e^{(t)} ) \big) \\ &\qquad \qquad \qquad \qquad + (1 - y_e) \log \big(1 - \sigma(e_e^{(t)}) \big) \Big)
    \end{split}
\end{equation}
where $y_{i,c}$ indicates if node $v_i$ belongs to the class $c$ (\ie $y_{i,c} = 1$ if $v_i$ belongs to class $c$, and $0$ otherwise), $\vp_{i,c}$ is the predicted probability of node $v_i$ belonging to class $c$, $e_e^{(t)}$ is the un-normalized attention score of edge $e$ in the $t$-th message passing layer, and $y_e$ is the label of the edge (\ie $1$ if source and target nodes belong to the same class, and $0$ otherwise).
An illustration of the proposed method is given in Figure~\ref{fig:architecture}.

\section{Experiments}\label{sec:exp}

\begin{figure}
\centering
\begin{subfigure}[b]{0.65\textwidth}{
   \includegraphics[width=\textwidth]{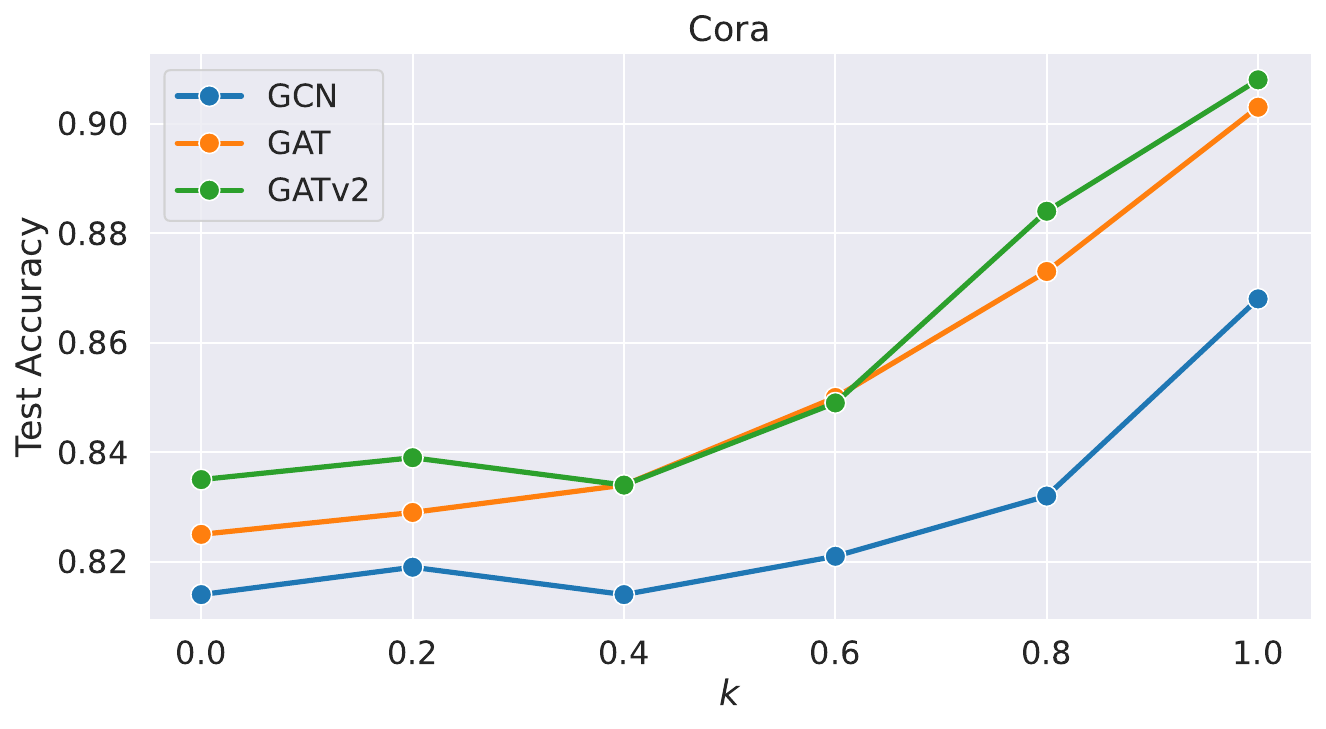}
   
}
\end{subfigure}
\vspace{-0.5cm}
\begin{subfigure}[b]{0.65\textwidth}{
   \includegraphics[width=\textwidth]{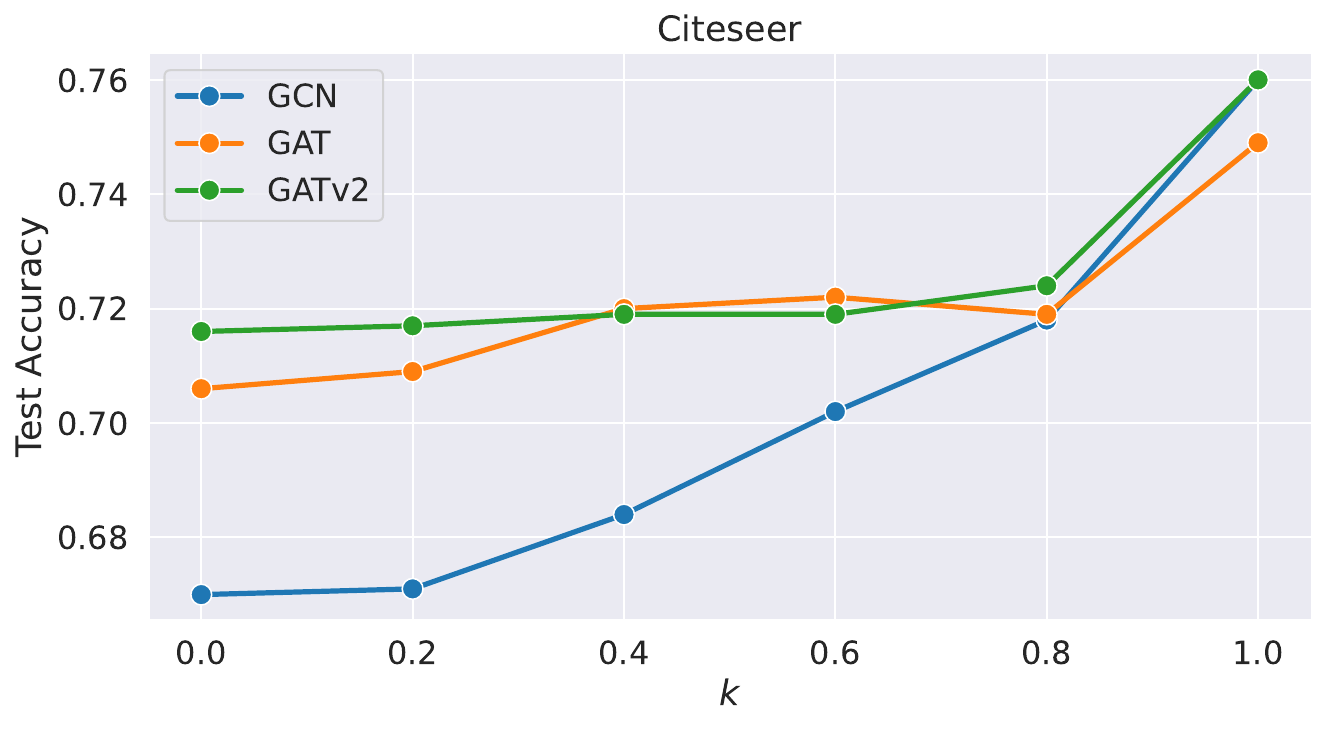}
}
\end{subfigure}
\vspace{-0.5cm}
\begin{subfigure}[b]{0.65\textwidth}{
   \includegraphics[width=\textwidth]{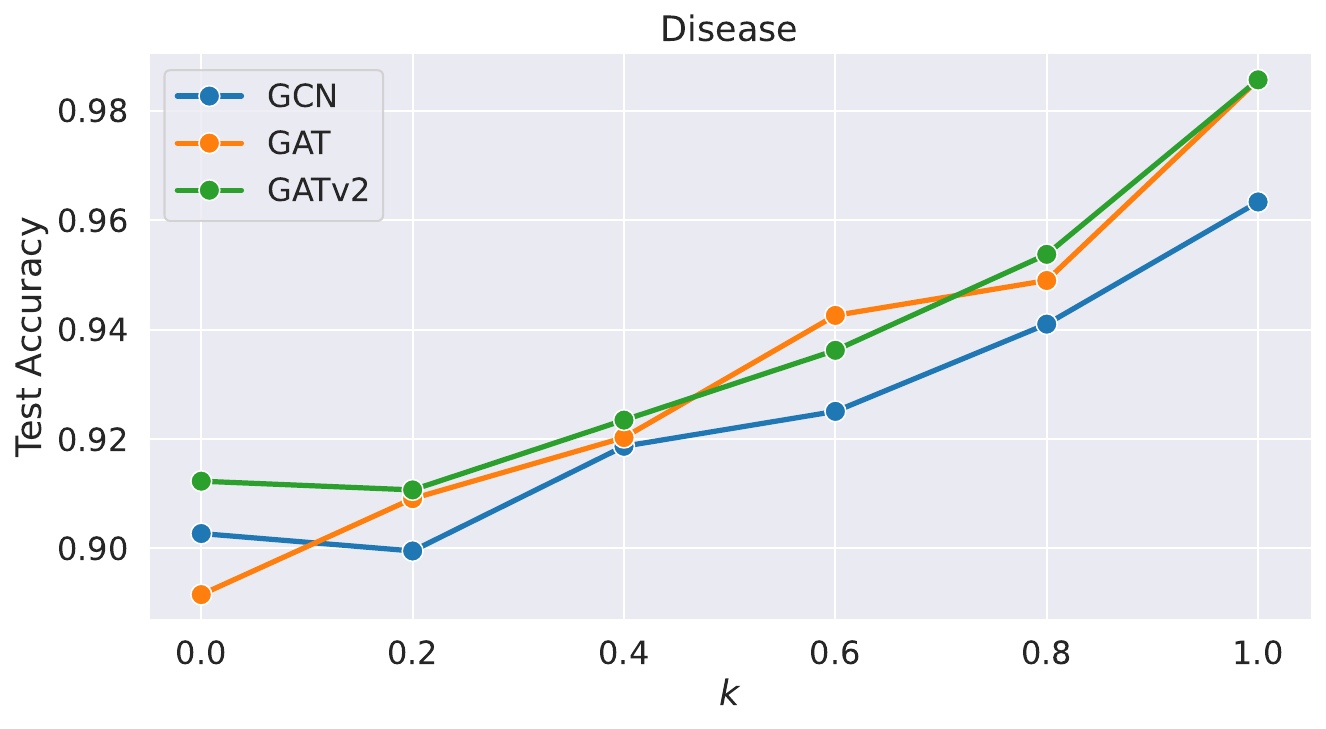}
} 
\end{subfigure}
\vspace{-0.5cm}
\begin{subfigure}[b]{0.65\textwidth}{
   \includegraphics[width=\textwidth]{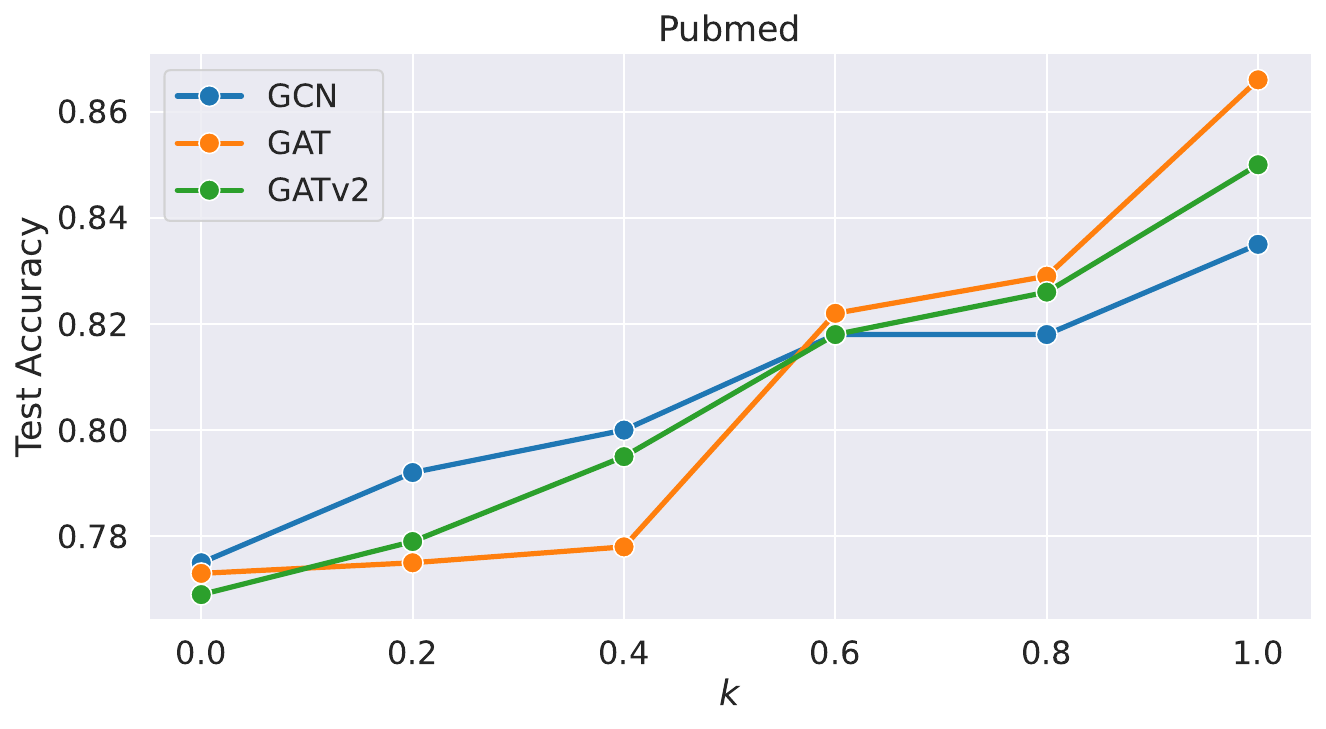}
}
\end{subfigure}
\vspace{-0.2cm}
\caption{Test performance of GNNs by removing $k|E_{\text{inter}}|$ inter-class edges. Setting $k = 0$ corresponds to the original graph and $k = 1$ corresponds to a fully homophilic graph. Performance improves as the ratio of homophilic edges increases.}
\label{exp:homophily}
\end{figure}

In this section, we extensively evaluate our method on synthetic as well as standard node classification datasets, and compare it against state-of-the-art GNNs.
As already mentioned, we apply the proposed method to the GATv2 model.
%Our code is available in the supplementary material.

\subsection{Adjusting Homophily in Graphs}
Our method is based on the assumption that node classification is easier in homophilic graphs, since nodes from the different classes will have separable representations. 
In this experiment, we try to verify this claim. 
Therefore, we test the performance of various GNNs by adjusting the edge homophily in various graph datasets.
Specifically, we remove $k|E_{\text{inter}}|$ inter-class edges from each dataset, where $|E_{\text{inter}}|$ is the number of inter-class edges in a graph.
Setting $k=0$ corresponds to the original graph and $k=1$ corresponds to a fully homophilic graph.  
We report the results for Cora, Citeseer and Disease datasets in Figure~\ref{exp:homophily}.
We observe that the performance increases for all the models, as the homophily of the graph increases. 
Our approach is strongly motivated by this observation, since the proposed loss function encourages the attention scores of inter-class edges to be close to $0$, thus generating a more homophilic-like setting.

\subsection{Node Classification Benchmarks}

\noindent\textbf{Baselines.} We compare our approach (HS-GATv2) against the following state-of-the-art GNN models: Graph Convolutional Network (GCN)~\cite{kipf2017semi}, GraphSAGE~\cite{hamilton2017inductive}, Graph Attention Network~(GAT)~\cite{velicko2018}, GATv2~\cite{brody2022how}, and Principal Neighbourhood Aggregation~(PNA)~\cite{NEURIPS2020_99cad265}.

\noindent\textbf{Datasets.}
We utilize four well-known node classification benchmark datasets to evaluate our approach in real-world scenarios. We use three citation network datasets: Cora, CiteSeer and Pubmed~\cite{sen2008collective}, where each node corresponds to a scientific publication, edges correspond to citations and the goal is to predict the category of each publication. 
We follow the experimental setup of~\cite{kipf2017semi} and use $140$ nodes for training, $300$ for validation and $1000$ for testing. 
We further use one disease spreading model: Disease~\cite{chami2019hyperbolic}. It simulates the SIR disease spreading model~\cite{nla.cat-vn2624541}, where the label of a node indicates if it is infected or not.
We follow the experimental setup of~\cite{chami2019hyperbolic} and use $30/10/60\%$ for training, validation and test sets and report the average results from $10$ different random splits.

\noindent\textbf{Experimental Setup.}
We use the Adam optimizer~\cite{kingma2014adam} with the Glorot initialization~\cite{glorot2010understanding}.
We search the layers from $\{1,2\}$ and the attention heads from $\{1,4,8\}$.
We set the weight decay equal to $\mathrm{5e}{-5}$.
We fix the mixing coefficient $\lambda$ to $0.1$.
We search the hidden dimensions from $\{8,16,32,64,128\}$, the learning rate from $\{0.001,0.005\}$ and the dropout rate from $\{0.0,0.2,0.5\}$.

\noindent\textbf{Results}
Table~\ref{results:node_benchmarks} illustrates the obtained test accuracies.
We observe that the proposed HS-GATv2 method outperforms the baselines on all three datasets.
This highlights the ability of the proposed approach to use the attention mechanism to reduce the noisy information that each node receives from its neighbors, thus producing high-quality node representations.

\begin{table}[t]
\centering
    \caption{Test accuracy in the node classification benchmarks.}
    \begin{tabular}{lcccc}
    \toprule
    \textbf{Method} & \textbf{Cora}  & \textbf{Citeseer} & 
    \textbf{Disease} & \textbf{Pubmed} \\ %\textbf{LastFM Asia} & \textbf{Computers} &
    %\textbf{Photo} & \\
    \midrule
    MLP & 43.8 & 52.9 & 79.1 $ \pm$ 1.0 & 74.2 $\pm$ 0.2\\ %& 72.27 $\pm 1.00$ & 79.53 $\pm 0.66$ & 87.89 $\pm 1.04$ \\
    % \hline
    GCN & 81.4 & 67.5 & 89.0 $\pm$ 2.2 & 77.8 $\pm$ 0.3\\ %& 83.58 $\pm 0.93$ & 90.72 $\pm 0.50$ &  93.99 $\pm 0.42$  \\
    GraphSAGE & 77.2 & 65.3 & 88.8 $\pm$ 2.0 & 77.9 $\pm$ 0.6 \\ %& 83.07 $\pm 1.19$ & 91.47 $\pm 0.37$ & 94.32 $\pm 0.46$ \\
    %GIN & 75.5 & 62.1 & 90.20 $\pm 2.23$ & 82.94 $\pm 1.25$ & 84.68 $\pm 2.33$ & 90.07 $\pm 1.19$ \\
    PNA & 76.4 & 58.9 & 86.8 $\pm$ 1.9 & 75.8 $\pm$ 0.6 \\ %83.24 $\pm 1.10$ & 90.80 $\pm 0.51$ & \underline{94.35} $\pm0.68$ & \\
    \hline
    GAT & 82.5 & 70.6 & 88.1 $\pm$ 2.5 & 78.1 $\pm$ 0.6 \\
    GATv2 & 83.5 & 71.6 & 89.2 $\pm$ 1.7 & 78.5 $\pm$ 0.4 \\
    \textbf{HS-GATv2 (ours)} & \textbf{85.3} & \textbf{73.5} & \textbf{89.3} $\pm$ 3.3 & \textbf{79.1} $\pm$ 0.3
     \\
    \bottomrule
    \end{tabular}
    \label{results:node_benchmarks}
\end{table}

%\begin{figure}[t]
%\centering
%  \includegraphics[width=1\columnwidth]{figures/deeper_cora.pdf}
%  \caption{Test performance of various GNNs with respect to different number of message passing layers in Cora.}
%  \label{fig:deep_cora}
%\end{figure}

\begin{figure}[t]
    \centering
    \includegraphics[width=0.8\columnwidth]{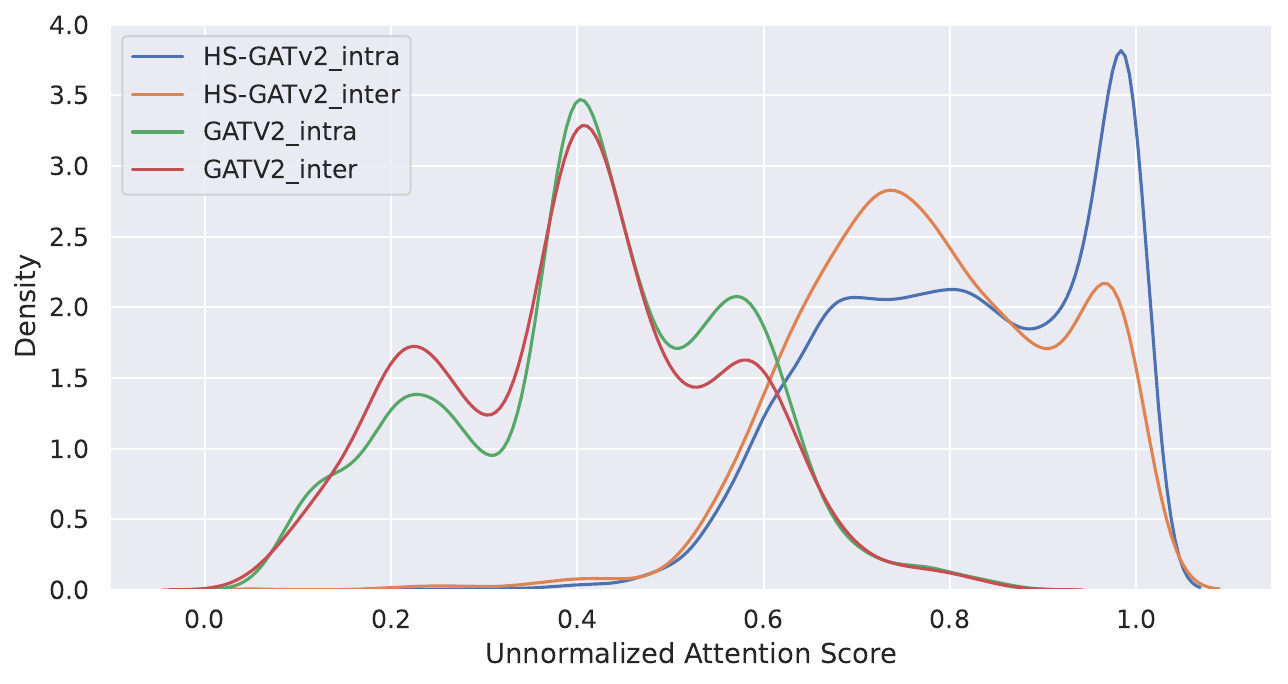}
    \caption{Distribution of attention scores averaged across the eight attention heads in Cora for intra and inter class edges that were not presented in the training set.}
    \label{fig:att_intra_iter}
\end{figure}

%\subsection{Increasing Depth in Graph Neural Networks}
%In this experiment, we examine the performance of our method in a deep setting, when multiple message passing layers are used.
%In Figure~\ref{fig:deep_cora}, we report the results for the Cora dataset.
%We observe that the performance of the proposed method does not decrease that much as the number of layers increases, and that it provides further improvements over the baselines in the deep scenario.
%\todo{check average value of attention scores between inter-intra edges}

\subsection{Distribution of Attention Scores}

In this experiment, we compute the distribution of the un-normalized attention scores produced by HS-GATv2 and GATv2 for edges whose endpoints are not in the training set. 
The results for the Cora dataset are illustrated in Figure~\ref{fig:att_intra_iter}. 
Attention scores obtained from GATv2 have the same distribution for the intra- and inter-class edges.
On the other hand, we observe that HS-GATv2 produces higher attention values for the intra-class edges even though it has not seen them during training.
This allows our model to reduce the noisy information in the message passing procedure, and to focus mainly on the homophilic edges.

\section{Conclusion}\label{sec:conclusions}
In this paper, we introduced a new type of graph attention model that uses supervision in the attention scores by exploiting the network homophily. 
Our proposed loss function contains a loss term that encourages attention scores to be high between nodes that share the same label and therefore alleviates the problem of information mixing in GNNs. 
Our extensive experiments demonstrate an increase in the performance of the proposed method over state-of-the-art GNNs such as GAT and GATv2 in node classification tasks.

\section*{Acknowledgements}
G.N. is supported by the French National research agency via the AML-HELAS (ANR-19-CHIA-0020) project.

\bibliographystyle{splncs04}
\bibliography{bib}

\begin{thebibliography}{10}
\providecommand{\url}[1]{\texttt{#1}}
\providecommand{\urlprefix}{URL }
\providecommand{\doi}[1]{https://doi.org/#1}

\bibitem{abu2019mixhop}
Abu-El-Haija, S., Perozzi, B., Kapoor, A., Alipourfard, N., Lerman, K.,
  Harutyunyan, H., Ver~Steeg, G., Galstyan, A.: {MixHop: Higher-Order Graph
  Convolutional Architectures via Sparsified Neighborhood Mixing}. In:
  Proceedings of the 36th International Conference on Machine Learning. pp.
  21--29 (2019)

\bibitem{nla.cat-vn2624541}
Anderson, R.M., May, R.M.: Infectious diseases of humans: dynamics and control.
  Oxford University Press (1992)

\bibitem{brody2022how}
Brody, S., Alon, U., Yahav, E.: How attentive are graph attention networks? In:
  10th International Conference on Learning Representations (2022)

\bibitem{cai2020note}
Cai, C., Wang, Y.: A note on over-smoothing for graph neural networks. arXiv
  preprint arXiv:2006.13318  (2020)

\bibitem{chami2019hyperbolic}
Chami, I., Ying, Z., R\'{e}, C., Leskovec, J.: Hyperbolic graph convolutional
  neural networks. In: Advances in Neural Information Processing Systems (2019)

\bibitem{chatzianastasis2023graph}
Chatzianastasis, M., Lutzeyer, J., Dasoulas, G., Vazirgiannis, M.: Graph
  ordering attention networks. In: Proceedings of the 37th AAAI Conference on
  Artificial Intelligence. pp. 7006--7014 (2023)

\bibitem{chen2020measuring}
Chen, D., Lin, Y., Li, W., Li, P., Zhou, J., Sun, X.: Measuring and relieving
  the over-smoothing problem for graph neural networks from the topological
  view. In: Proceedings of the 34th AAAI conference on artificial intelligence.
  pp. 3438--3445 (2020)

\bibitem{NEURIPS2020_99cad265}
Corso, G., Cavalleri, L., Beaini, D., Li\`{o}, P., Veli\v{c}kovi\'{c}, P.:
  Principal neighbourhood aggregation for graph nets. In: Advances in Neural
  Information Processing Systems. pp. 13260--13271 (2020)

\bibitem{NEURIPS2020_c9f2f917}
Cranmer, M., Sanchez~Gonzalez, A., Battaglia, P., Xu, R., Cranmer, K., Spergel,
  D., Ho, S.: Discovering symbolic models from deep learning with inductive
  biases. In: Advances in Neural Information Processing Systems. pp.
  17429--17442 (2020)

\bibitem{dasoulas2021learning}
Dasoulas, G., Lutzeyer, J., Vazirgiannis, M.: {Learning Parametrised Graph
  Shift Operators}. In: 9th International Conference on Learning
  Representations (2021)

\bibitem{dasoulas2021lipschitz}
Dasoulas, G., Scaman, K., Virmaux, A.: Lipschitz normalization for
  self-attention layers with application to graph neural networks. In:
  Proceedings of the 38th International Conference on Machine Learning. pp.
  2456--2466 (2021)

\bibitem{gilmer2017neural}
Gilmer, J., Schoenholz, S.S., Riley, P.F., Vinyals, O., Dahl, G.E.: Neural
  message passing for quantum chemistry. In: Proceedings of the 34th
  International conference on machine learning. pp. 1263--1272 (2017)

\bibitem{glorot2010understanding}
Glorot, X., Bengio, Y.: Understanding the difficulty of training deep
  feedforward neural networks. In: Proceedings of the 13th International
  Conference on Artificial Intelligence and Statistics. pp. 249--256 (2010)

\bibitem{gori2005new}
Gori, M., Monfardini, G., Scarselli, F.: A new model for learning in graph
  domains. In: Proceedings of the 2005 IEEE International Joint Conference on
  Neural Networks. vol.~2, pp. 729--734 (2005)

\bibitem{hamilton2017inductive}
Hamilton, W., Ying, Z., Leskovec, J.: {Inductive Representation Learning on
  Large Graphs}. In: Advances in Neural Information Processing Systems. pp.
  1024--1034 (2017)

\bibitem{han2020graph}
Han, Y., Karunasekera, S., Leckie, C.: Graph neural networks with continual
  learning for fake news detection from social media. arXiv preprint
  arXiv:2007.03316  (2020)

\bibitem{jin2020gralsp}
Jin, Y., Song, G., Shi, C.: {GraLSP: Graph Neural Networks with Local
  Structural Patterns}. In: Proceedings of the 34th AAAI Conference on
  Artificial Intelligence. pp. 4361--4368 (2020)

\bibitem{karhadkar2022fosr}
Karhadkar, K., Banerjee, P.K., Montufar, G.: Fosr: First-order spectral
  rewiring for addressing oversquashing in gnns. In: 11th International
  Conference on Learning Representations (2022)

\bibitem{kim2021how}
Kim, D., Oh, A.: How to find your friendly neighborhood: Graph attention design
  with self-supervision. In: International Conference on Learning
  Representations (2021)

\bibitem{kingma2014adam}
Kingma, D.P., Ba, J.: Adam: A method for stochastic optimization. In: 3rd
  International Conference on Learning Representations (2014)

\bibitem{kipf2017semi}
Kipf, T.N., Welling, M.: {Semi-Supervised Classification with Graph
  Convolutional Networks}. In: 5th International Conference on Learning
  Representations (2017)

\bibitem{li2022graph}
Li, M.M., Huang, K., Zitnik, M.: Graph representation learning in biomedicine
  and healthcare. Nature Biomedical Engineering  \textbf{6}(12),  1353--1369
  (2022)

\bibitem{liu2019hyperbolic}
Liu, Q., Nickel, M., Kiela, D.: Hyperbolic graph neural networks. In: Advances
  in Neural Information Processing Systems. pp. 8230--8241 (2019)

\bibitem{mahmood2021masked}
Mahmood, O., Mansimov, E., Bonneau, R., Cho, K.: Masked graph modeling for
  molecule generation. Nature communications  \textbf{12}(1),  1--12 (2021)

\bibitem{min2020scattering}
Min, Y., Wenkel, F., Wolf, G.: Scattering gcn: Overcoming oversmoothness in
  graph convolutional networks. In: Advances in Neural Information Processing
  Systems. pp. 14498--14508 (2020)

\bibitem{murphy2019relational}
Murphy, R., Srinivasan, B., Rao, V., Ribeiro, B.: {Relational Pooling for Graph
  Representations}. In: Proceedings of the 36th International Conference on
  Machine Learning. pp. 4663--4673 (2019)

\bibitem{nikolentzos2023weisfeiler}
Nikolentzos, G., Chatzianastasis, M., Vazirgiannis, M.: Weisfeiler and leman go
  hyperbolic: Learning distance preserving node representations. In:
  Proceedings of the 26th International Conference on Artificial Intelligence
  and Statistics. pp. 1037--1054 (2023)

\bibitem{nikolentzos2020k}
Nikolentzos, G., Dasoulas, G., Vazirgiannis, M.: k-hop graph neural networks.
  Neural Networks  \textbf{130},  195--205 (2020)

\bibitem{nikolentzos2020message}
Nikolentzos, G., Tixier, A., Vazirgiannis, M.: Message passing attention
  networks for document understanding. In: Proceedings of the 34th AAAI
  Conference on Artificial Intelligence. pp. 8544--8551 (2020)

\bibitem{scarselli2008graph}
Scarselli, F., Gori, M., Tsoi, A.C., Hagenbuchner, M., Monfardini, G.: The
  graph neural network model. IEEE transactions on neural networks
  \textbf{20}(1),  61--80 (2008)

\bibitem{sen2008collective}
Sen, P., Namata, G., Bilgic, M., Getoor, L., Galligher, B., Eliassi-Rad, T.:
  Collective classification in network data. AI magazine  \textbf{29}(3),
  93--93 (2008)

\bibitem{seo2019discriminative}
Seo, Y., Loukas, A., Perraudin, N.: Discriminative structural graph
  classification. arXiv preprint arXiv:1905.13422  (2019)

\bibitem{sperduti1997supervised}
Sperduti, A., Starita, A.: {Supervised Neural Networks for the Classification
  of Structures}. IEEE Transactions on Neural Networks  \textbf{8}(3),
  714--735 (1997)

\bibitem{tsubaki2019compound}
Tsubaki, M., Tomii, K., Sese, J.: Compound--protein interaction prediction with
  end-to-end learning of neural networks for graphs and sequences.
  Bioinformatics  \textbf{35}(2),  309--318 (2019)

\bibitem{velicko2018}
Veli{\v{c}}kovi{\'c}, P., Cucurull, G., Casanova, A., Romero, A., Lio, P.,
  Bengio, Y.: Graph attention networks. In: 6th International Conference on
  Learning Representations (2018)

\bibitem{wang2019improving}
Wang, G., Ying, R., Huang, J., Leskovec, J.: Improving graph attention networks
  with large margin-based constraints. arXiv preprint arXiv:1910.11945  (2019)

\bibitem{xu2019powerful}
Xu, K., Hu, W., Leskovec, J., Jegelka, S.: {How Powerful are Graph Neural
  Networks?} In: 7th International Conference on Learning Representations
  (2019)

\bibitem{yang2020revisiting}
Yang, C., Wang, R., Yao, S., Liu, S., Abdelzaher, T.: Revisiting over-smoothing
  in deep gcns. arXiv preprint arXiv:2003.13663  (2020)

\bibitem{Zhao2020PairNorm}
Zhao, L., Akoglu, L.: Pairnorm: Tackling oversmoothing in gnns. In: 8th
  International Conference on Learning Representations (2020)

\end{thebibliography}

\end{document}